\documentclass{article}

\usepackage{microtype}
\usepackage{graphicx}
\usepackage{subfigure}
\usepackage{booktabs} 

\usepackage{hyperref}

\usepackage[accepted]{icml2024}

\usepackage{amsmath}
\usepackage{amssymb}
\usepackage{mathtools}
\usepackage{amsthm}

\usepackage{bm}
\usepackage{graphicx}
\usepackage{arydshln}
\usepackage{bbding}
\usepackage{array}
\newcolumntype{C}[1]{>{\centering\let\newline\\\arraybackslash\hspace{0pt}}m{#1}}
\usepackage{multirow}
\usepackage[capitalize,noabbrev]{cleveref}

\theoremstyle{plain}

\theoremstyle{definition}

\theoremstyle{remark}

\usepackage[textsize=tiny]{todonotes}

\icmltitlerunning{SpikeLM: Towards General Spike-Driven Language Modeling}

\begin{document}

\twocolumn[
\icmltitle{SpikeLM: Towards General Spike-Driven Language Modeling \\ 
            via Elastic Bi-Spiking Mechanisms}



\icmlsetsymbol{equal}{*}

\begin{icmlauthorlist}
\icmlauthor{Xingrun Xing}{a,b,c}
\icmlauthor{Zheng Zhang}{c}
\icmlauthor{Ziyi Ni}{a,b}
\icmlauthor{Shitao Xiao}{c}
\icmlauthor{Yiming Ju}{c}
\icmlauthor{Siqi Fan}{c}
\icmlauthor{Yequan Wang}{c}
\icmlauthor{Jiajun Zhang}{a,b}
\icmlauthor{Guoqi Li}{a}
\end{icmlauthorlist}



\icmlaffiliation{a}{Institute of Automation, Chinese Academy of Sciences}
\icmlaffiliation{b}{School of Artificial Intelligence, University of Chinese Academy of Sciences}
\icmlaffiliation{c}{Beijing Academy of Artificial Intelligence}

\icmlcorrespondingauthor{Zheng Zhang}{zhangz.goal@gmail.com}
\icmlcorrespondingauthor{Jiajun Zhang}{jjzhang@nlpr.ia.ac.cn}
\icmlcorrespondingauthor{Guoqi Li}{guoqi.li@ia.ac.cn}

\icmlkeywords{Spiking Neural Network, Language Model, Energy Efficiency}

\vskip 0.3in
]



\printAffiliationsAndNotice{}  

\begin{abstract}
Towards energy-efficient artificial intelligence similar to the human brain, the bio-inspired spiking neural networks (SNNs) have advantages of biological plausibility, event-driven sparsity, and binary activation. 
Recently, large-scale language models exhibit promising generalization capability, making it a valuable issue to explore more general spike-driven models.
However, the binary spikes in existing SNNs fail to encode adequate semantic information, placing technological challenges for generalization.
This work proposes the first fully spiking mechanism for general language tasks, including both discriminative and generative ones.
Different from previous spikes with \{0,1\} levels, we propose a more general spike formulation with bi-directional, elastic amplitude, and elastic frequency encoding, while still maintaining the addition nature of SNNs. 
In a single time step, the spike is enhanced by direction and amplitude information; in spike frequency, a strategy to control spike firing rate is well designed.
We plug this elastic bi-spiking mechanism in language modeling, named SpikeLM. 
It is the first time to handle general language tasks with fully spike-driven models, which achieve much higher accuracy than previously possible.
SpikeLM also greatly bridges the performance gap between SNNs and ANNs in language modeling.
Our code is available at https://github.com/Xingrun-Xing/SpikeLM.
\end{abstract}

\section{Introduction}

Creating artificial general intelligence by simulating the human brain has always been a human dream, which is known as Brain-Inspired Computing (BIC) \cite{mehonic2022brain,zhang2020system}. 
Although artificial neural networks (ANNs) \cite{touvron2023llama, kirillov2023segment} have achieved tremendous success, the working ways are still so different from the human brain. 
The biological neurons communicate with spikes \cite{roy2019towards} and only activate when the membrane potential exceeds a certain threshold. 
Spiking neural networks (SNNs) \cite{maass1997networks} are designed by simulating biological neuron dynamics \cite{gerstner2014neuronal}, and have distinctive attributions of biological plausiblity, event-driven sparsity, and binary activation. 
Given event-driven computation, high sparsity is dynamically achieved by event occurrence.
Given binary activations, matrix multiplications convert to accumulate (AC) operations.
These characteristics make bio-inspired SNNs a significantly energy-efficient alternative \cite{yin2021accurate,schuman2022opportunities} to traditional ANNs.
Deepening our understanding of spiking neurons \cite{fang2021incorporating} and expanding the usage scope of SNNs \cite{kim2020spiking,zhou2024spikformer} have become increasingly valuable issues. 

Previous SNNs mainly focus on computer vision \cite{wu2021progressive, hu2021spiking} due to relatively simple tasks and smaller model sizes.
Recently, large-scale language models \cite{touvron2023llama, du2022glam} exhibit much more advanced generalization ability \cite{brown2020language} than other fields in machine learning, which motivates us to pursue more general spike-driven models with language modeling.
However, this objective is no-trial due to the technological challenges in spike representation \cite{deng2020optimal, guo2023rmp} and optimization \cite{neftci2019surrogate, wu2018spatio, guo2023joint}.
In representation, binary spike leads to severe information loss \cite{deng2020optimal}, making it difficult to generalize across language tasks.
In optimization, large-scale language models require stable and highly efficient gradient calculation, while neuronal dynamics in SNNs are non-differentiable.
Therefore, there are very limited language-oriented SNNs.

This work focuses on fully spike-driven language modeling in general tasks, including both discriminative and generative ones, which is not addressed in the previous SNN studies. 
Notably, fully spike-driven indicates replacing all matrix multiplications as spike operations, except the last regression.
To explore the capabilities and limitations of current SNNs, we initially apply existing SNN technologies \cite{hu2021spiking,gerstner2014neuronal} to construct fully spike-driven baselines. 
Basically, there is a large performance gap between language-oriented ANNs and SNNs. 
Moreover, the fixed spike firing rate in SNNs makes a suboptimal trade-off between performance and energy efficiency.

\begin{figure}[t]
\begin{center}
\centerline{\includegraphics[width=0.97\columnwidth]{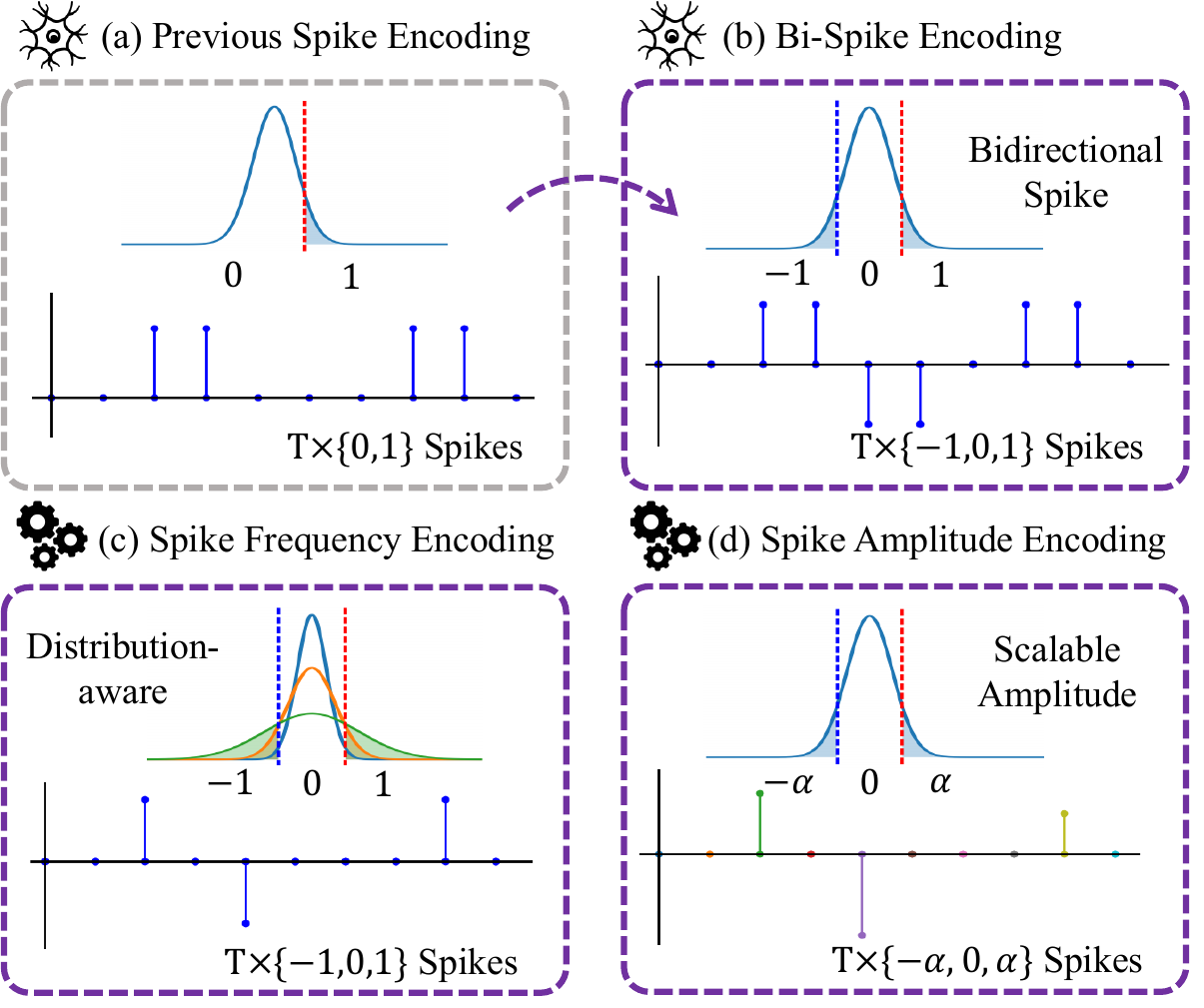}}
\caption{Comparisons between previous spike encoding (a) and our elastic bidirectional encodings (b, c, d). The bidirectional, frequency and amplitude encodings are sequentially applied .}
\label{f1}
\end{center}
\vskip -0.3in
\end{figure}

To address the aforementioned issues, we focus on boosting modeling capabilities of SNNs through generalized spike encoding methods. To extend semantic information, we sequentially generalize spike formulations as shown in Fig.\ref{f1}: \\
\textbf{(i) Bi-directional spike encoding.} Different from previous binary spike levels $\{0,1\}$, we propose bidirectional spikes with ternary levels $\{-1,0,1\}$. Bidirectional encoding doubles semantic information and maintains the addition nature of SNNs at the same time. \\
\textbf{(ii) Elastic spike frequency encoding.} Different from previous empirical spike firing rates, we encode spike frequency according to input distributions, achieving a controllable firing rate for better performance and energy trade-off. \\
\textbf{(iii) Elastic spike amplitude encoding.} To retain membrane potential intensity, we encode spike with amplitude information as $\{-\alpha,0,\alpha\}$. A layerwise $\alpha$ is used, which can be merged with weights after training. Therefore, the addition nature of SNNs is still maintained. \\
Given a multi-step spike, these encoding methods jointly extend spike capabilities by direction and amplitude in each time step and frequency across time steps.
We plug this elastic bi-spiking mechanism in language modeling, termed SpikeLM. 
Thanks to improved spikes, as Table \ref{fully}, it achieves the first fully spiking mechanism in general language tasks, by replacing all matrix multiplications in ANNs.
Our contributions are summarised as follows:
\vspace{-0.2cm}
\begin{itemize}
\item We propose SpikeLM, the first general fully spike-driven language modeling, significantly broadening the usage scope of language-oriented SNNs. As Table \ref{res}, SpikeLM achieves much higher accuracy than what was possible previously and largely bridges the performance gap between SNNs and ANNs.
\item We propose an elastic bi-spiking mechanism. At the same time, it maintains the addition nature of SNNs. A controllable spike firing rate is also achieved.
\item We introduce the dynamic isometry \cite{chen2020comprehensive}, and theoretically prove that the training stability of the elastic bi-spiking function surpasses the ReLU function in ANNs, ensuring stable optimization for SpikeLMs.
\end{itemize}

\begin{table}[t]
\caption{Comparisons with SpikeBERT \cite{lv2023spikebert} and SpikeGPT \cite{zhu2023spikegpt}. The "+" sign indicates the level of capability, with SpikeGPT only utilized for basic language modeling, lacking applications in sentence-level generation tasks. AC and MAC indicate ACcumulate and Multiply-ACcumulate.}
\label{fully}
\begin{center}
\begin{small}
\begin{tabular}{l|ccc}
\toprule
Task/Operation    & SpikeBERT & SpikeGPT & SpikeLM \\
\midrule
Discrimination    & \texttt{\textbf{+++}}     & \texttt{\textbf{+++}}      & \texttt{\textbf{+++}} \\
Generation        & \texttt{\textbf{-\phantom{++}}}         & \texttt{\textbf{+\phantom{++}}}        & \texttt{\textbf{+++}} \\
\midrule
Spike-Driven      & Fully      & Partly    & Fully \\
Matrix Mul.       & ACs      & MACs    & ACs \\
\bottomrule
\end{tabular}
\end{small}
\end{center}
\vskip -0.2in
\end{table}

\vspace{-0.3cm}
\begin{table}[t]
\caption{Comparisons between ANN and SNNs in generative and discriminative tasks in the same BERT or BART architecture.}
\label{res}
\begin{center}
\begin{small}
\begin{tabular}{l|c|cc}
\toprule
Dataset & ANN & LIF-SNN & SpikeLM \\
\midrule
MNLI$_\texttt{-m/mm}$ (Acc.)  & 83.8/83.4 & 56.8/55.2 & 77.1/77.2 \\
MRPC (F1)    & 89.8      & 82.3      & 85.7 \\
STS-2 (Acc)  & 92.3      & 80.6       & 87.0 \\
RTE (Acc.)   & 69.3      & 53.8      & 69.0 \\
STS-B (SP.)  & 89.4      & 20.0       & 84.9 \\
\midrule
XSUM (R-L)   & 34.7      & 28.3      & 32.9 \\
CNN-DM (R-L) & 31.7      & 28.1      & 29.1 \\
WMT16 (BLEU) & 26.8      & 19.0       & 23.0 \\
\midrule
Average      & 66.8      & 47.1      & 62.9 \\
\bottomrule
\end{tabular}
\end{small}
\end{center}
\vskip -0.2in
\end{table}

\section{Related Work}
\textbf{Bio-inspired SNNs.} BIC field \cite{mehonic2022brain,zhang2020system} is boosted by both advanced neuroscience and deep learning. 
Recent BIC field gets inspiration from learning rules \cite{payeur2021burst}, structures \cite{pham2021dualnet}, and energy-efficient computation \cite{schuman2022opportunities, yao2023sparser} in the nervous system. 
As a BIC algorithm, SNN  \cite{maass1997networks} also takes advantages of deep learning, for example, the spike-driven residual learning \cite{fang2021deep}, normalization \cite{zheng2021going, guo2023membrane}, self-attention \cite{yao2023spike,zhou2023spikingformer, yao2024spike}, backpropagation \cite{meng2022training, li2021differentiable, su2023deep, guo2023direct}, and ANN-SNN conversion \cite{bu2021optimal,deng2020optimal} technologies. One of the recent works also introduces ternary spikes \cite{guo2024ternary} in the computer vision field, while this work is in parallel with it and has different frequency and amplitude encoding to reduce average firing rate in language tasks.
Inspired by generalization capability in both the spike-driven human brain \cite{gerstner2014neuronal,izhikevich2003simple} and recent large language models \cite{touvron2023llama,brown2020language}, we are the first time to explore fully spike-driven models in general language tasks. 

\textbf{Neuromorphic chips.} Neuromorphic chips are inspired by the brain with non-von Neumann architectures \cite{Speck,schuman2022opportunities,roy2019towards,merolla2014million}. Owning the high sparsity and event-driven SNNs, their energy consumption can be tens to hundreds of mWs \cite{basu2022spiking} in SNNs workloads by compute gating or clock gating techniques \cite{narayanan2020spinalflow}.

\textbf{Language-oriented SNNs.} SpikeBERT \cite{lv2023spikebert} distills the spikingformer \cite{zhou2023spikingformer} in some discriminative tasks. However, performance drops to 59.7\% on the GLUE. 
SpikeGPT \cite{zhu2023spikegpt} introduces spike propagation between transformer blocks, but overall blocks are still ANNs. 
Compared with recent weight-quantized language models, BitNet \cite{wang2023bitnet} and ternary BitNet \cite{ma2024era}, SNNs more concentrate on bio-plausible activation spike encoding.

\section{Problem Formulation}
\subsection{Language Modeling with Vanilla SNNs}
We start by developing the first general baseline for fully spike-driven language modeling. 
Without loss of generality, we apply the most popular Leaky Integrate-and-Fire (LIF) neurons \cite{gerstner2014neuronal} to encode real-valued activations into spike sequences. 
Fully spike-driven transformers are achieved through LIF neurons in linear layers and the key and value of self-attention \cite{vaswani2017attention}.

\textbf{Spike encoding in linear.}
LIF neurons are neuronal dynamics added before linear layers, which output binary spikes with $\{0,1\}$ levels. The following matrix multiplication is converted to additions.
By simulating the charging and firing of biological neurons, LIF neurons can be governed by:
\begin{equation}
\label{spike1}
\bm{m}^l(t)= \bm{v}^l(t-1)+\bm{x}^{l-1}(t),
\end{equation}
\vspace{-0.2cm}
\begin{equation}
\label{spike2}
\bm{s}^{l}(t)= \begin{cases}0, & \text { if } \bm{m}^{l}(t) < \bm{\theta}^{l} \\ 1, & \text { if } \bm{m}^{l}(t) \ge \bm{\theta}^{l} \end{cases}, 
\end{equation}
\begin{equation}
\label{spike3}
\bm{v}^l(t)={\beta}\bm{m}^l(t)(1 - \bm{s}^{l}(t)) + {v}_{reset}\bm{s}^{l}(t).
\end{equation}
At each time step $t$, the LIF neuron performs a spike encoding until a certain spike length $T$.
$\bm{m}^l(t)$ and $\bm{v}^l(t)$ indicate the membrane potential before and after spike encoding respectively. 
To simulate the charging process, $\bm{m}^l(t)$ adds the inputs $\bm{x}^{l-1}(t)$ at the current moment to the membrane potential $\bm{v}^l(t-1)$ from the last moment. 
When the membrane potential $\bm{m}^l(t)$ exceeds the firing threshold $\bm{\theta}^{l}$, the neuron is triggered and the spike $\bm{s}^{l}(t)$ is encoded as 1; otherwise, it is 0. 
After spike encoding, the membrane potential $\bm{v}^l(t)$ is reset to a certain potential ${v}_{reset}$ if the spike is 1; otherwise, it will decay by a factor $\beta (< 1)$.

\textbf{Spike encoding in the key and value.}
The matrix multiplications in self-attention include the multiplication between the key and query, and the multiplication between the attention map and value.
By encoding the key and value as spikes by Eq.\ref{spike1},\ref{spike2},\ref{spike3}, all matrix multiplications are converted to additions. We set $\beta = 0$ in key and value. Notably, the time step of SNNs is an additional dimension.

\begin{figure}[t]
\begin{center}
\centerline{\includegraphics[width=\columnwidth]{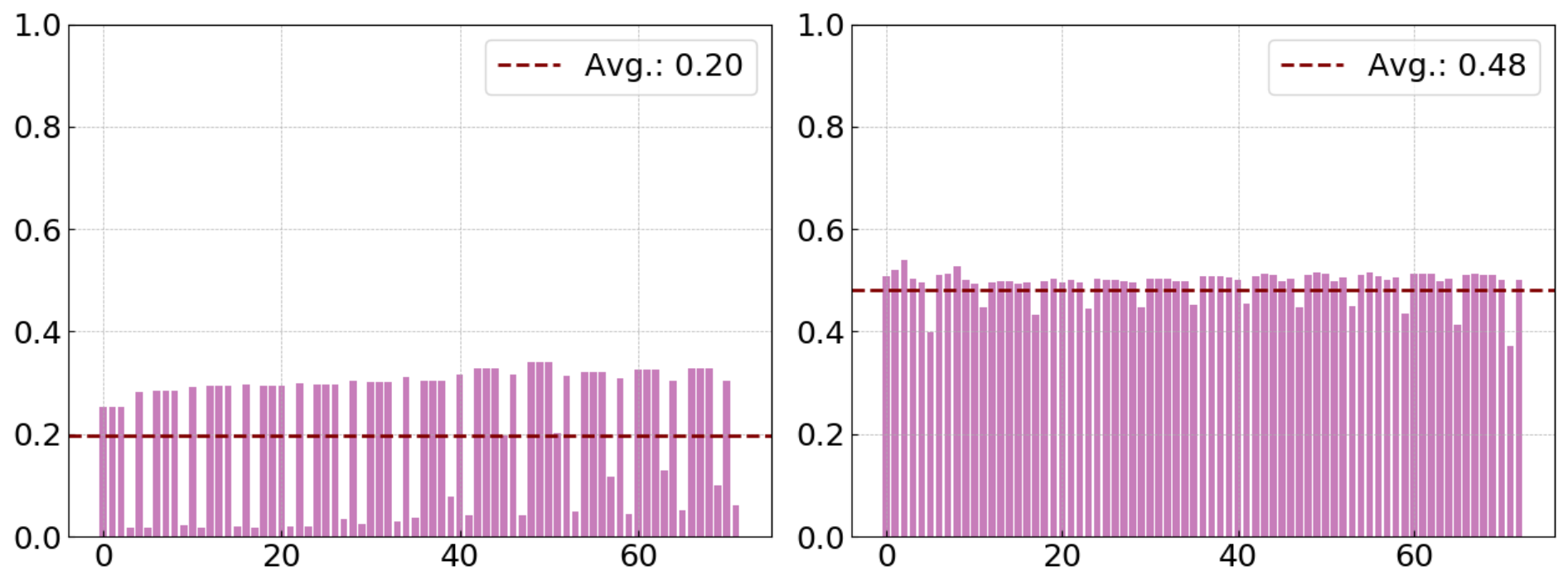}}
\caption{Spike firing rate in LIF-BERT (left) and activated rate in Binary BERT (right) in every linear layer.}
\label{fire0}
\end{center}
\vskip -0.55in
\end{figure}

We construct LIF-based transformers for both BERT \cite{devlin2018bert} and BART \cite{lewis2019bart} architectures, termed LIF-BERT and LIF-BART, for discriminative and generative tasks respectively. For optimization, we propose a straight-through estimator (STE) \cite{bengio2013estimating} based backpropagation in Appendix A.1, which achieves a strong baseline in general language tasks.

\subsection{Performance \& Energy Efficiency in SNNs}
Previous SNNs directly encode spikes, leading to an ill-posed problem: 
when the spike firing rate is low, it leads to a reduced information entropy in the Bernoulli-distributed spikes, limiting model capability. 
When the spike firing rate is high, it decreases the sparsity of the spikes, resulting in increased energy consumption. 

\textbf{Performance drop.}
As shown in Table \ref{res}, compared with ANNs, the LIF-based SNNs are driven by sparse binary spike, resulting in the average 19.7\% performance drop. We analyze the spike firing rate in LIF-BERT in Fig.\ref{fire0}.
Although the LIF-BERT has a low firing rate, the sparse and binary encoded spike is too simple without much capability to represent semantic information.

\textbf{Energy efficiency.}
We consider a case of the high firing rate. 
We directly replace the LIF neuron with a binary quantization function for one-step spike encoding following BiPFT \cite{xing2024bipft}, which is a Binary BERT with the \{-1, +1\} binarization level. For binary neural networks (BNNs) \cite{courbariaux2016binarized, xing2022towards}, the binary activations can map to \{0, 1\} in inference. As shown in Fig.\ref{fire0}, the proportion of 1 in Binary-BERT is close to 50\%. Compared with SNNs, BNNs demonstrate a much higher activation rate. 
With equally probable 0 and 1, information entropy in the Bernoulli distribution approaches maximum. However, the reduced sparsity leads to significantly increased energy consumption.

\section{General Spike Language Modeling}
We divide and conquer the problem of effective spike encoding into three aspects: spike direction encoding, spike frequency encoding, and spike amplitude encoding. 
From these perspectives, we present general and advanced spike encoding strategies, significantly enhancing the overall representational capacity of the spike signals.
Finally, we theoretically confirm the effectiveness of our spike encoding methods in general language-oriented SNNs.

\subsection{Bi-Directional Spike Encoding}
As shown in Eq.\ref{spike2}, the previous spike encoding binarizes the current membrane potential into $\{0,1\}$, overlooking all negative membrane potentials with half of the information. 
We propose a bidirectional spike encoding with ternary levels $\{-1,0,1\}$, considering both positive and negative membrane potentials. 
Since the spike encoding of the membrane potential is non-differentiable, we first define stochastic spike encoding to relax spikes as random variables. Then, we calculate the expectation of gradient based on the distribution of the spikes for backward propagation.

We first define positive stochastic spikes $\widetilde{\bm{s}}^{+}(t)$ and negative stochastic spikes $\widetilde{\bm{s}}^{-}(t)$ to encode positive and negative membrane potentials respectively. And then, the bidirectional stochastic spike $\widetilde{\bm{s}}^{\pm}(t)$ can be defined as the summarization of positive and negative spikes:
\begin{equation}
\label{ternary1}
\widetilde{\bm{s}}^{+}(t) \stackrel{\mathrm{def}}{=} \begin{cases}0, & \bm{p}^{0} = {\rm clip}(1 - \bm{m}(t), 0, 1) \\ +1, & \bm{p}^{+} =  {\rm clip}(\bm{m}(t), 0, 1) \end{cases}, 
\end{equation}
\begin{equation}
\label{ternary2}
\widetilde{\bm{s}}^{-}(t) \stackrel{\mathrm{def}}{=} \begin{cases}0, & \bm{p}^{0} = {\rm clip}(1 + \bm{m}(t), 0, 1) \\ -1, & \bm{p}^{-} =  {\rm clip}(-\bm{m}(t), 0, 1) \end{cases}, 
\end{equation}
\begin{equation}
\label{ternary3}
\widetilde{\bm{s}}^{\pm}(t) \stackrel{\mathrm{def}}{=} \widetilde{\bm{s}}^{+}(t) + \widetilde{\bm{s}}^{-}(t), 
\end{equation}
where $\bm{p}^{+}$, $\bm{p}^{0}$, and $\bm{p}^{-}$ indicate the probability of +1, 0, and -1 respectively. We define the $\bm{p}^{+}$, $\bm{p}^{0}$, and $\bm{p}^{-}$ according to their distance to the value of +1, 0 and -1, and the ${\rm clip}(.)$ operations confirm the probability in the range of [0,1], so that, the definitions of $\widetilde{\bm{s}}^{+}(t)$ and $\widetilde{\bm{s}}^{-}(t)$ confirm $\bm{p}^{+} + \bm{p}^{0} = 1$ and $\bm{p}^{-} + \bm{p}^{0} = 1$ respectively.

\textbf{Backward propagation.}
Eq.\ref{ternary1} and \ref{ternary2} are non-differentiable. To enable backpropagation in the entire SNN, we calculate the expectation of the stochastic gradient of $\widetilde{\bm{s}}^{\pm}(t)$. We use the gradient expectation $\mathbb{E}_{\widetilde{\bm{s}}^{\pm}(t)}$ in place of the deterministic gradient to complete the backpropagation:
\begin{equation}
\label{back}
\begin{split}
& \mathbb{E}_{\widetilde{\bm{s}}^{\pm}(t)}[\frac{\partial \widetilde{\bm{s}}^{\pm}(t)}{\partial \bm{m}(t)}] = \frac{\partial}{\partial \bm{m}(t)} \mathbb{E}[\widetilde{\bm{s}}^{\pm}(t)] \\
& \ \ = \frac{\partial}{\partial \bm{m}(t)} (-1 \times \bm{p}^{-} + 0 \times \bm{p}^{0} + 1 \times \bm{p}^{+}) \\
& \ \ = \frac{\partial}{\partial \bm{m}(t)} {\rm clip}(\bm{m}(t),-1,1) \\
\end{split},
\end{equation}
where the gradient expectation $\mathbb{E}_{\widetilde{\bm{s}}^{\pm}(t)}$ can be derived as the straight-through estimator (STE) \cite{bengio2013estimating}, which is widely applied to relax non-differentiable operations. The backpropagation can achieve high efficiency, which only performs gradient identity between +1 and -1.

\textbf{Forward propagation.}
Eq.4 and 5 involve random sampling. In practice, we convert stochastic spike encoding into deterministic by setting fixed thresholds for efficiency:
\begin{equation}
\label{certain1}
{\bm{s}}^{\pm}(t) = \begin{cases} -1, & \text { if } \bm{m}(t)<-1 \\ 0, & \text { if } \bm{m}(t) \in (-1, +1) \\ +1, & \text { if } \bm{m}(t) > +1 \end{cases}, 
\end{equation}
which is derived by setting the probability condition $\bm{p}^{+}=1$ in Eq.\ref{ternary1} and $\bm{p}^{-}=1$ in Eq.\ref{ternary2} for +1 and -1 spike encoding respectively. After bidirectional spike encoding, the membrane potentials are encoded by sequences of \{+1,0,-1\}. 
Notably, matrix multiplications between bidirectional spikes and real-valued weights can be converted to pure addition and subtraction operations.

Under the same firing rate $r$, a bidirectional spike can, at most, increases information entropy $r$ bits for each time compared to the unidirectional spike.
This is achieved by directly calculating $\mathcal{H}({\bm{s}}_i^{\pm}(t)) - \mathcal{H}({\bm{s}}_i^{+}(t))$, where the information entropy of the original and bidirectional spikes are formulated by Eq.\ref{h1}:
\begin{equation}
\label{h1}
\begin{split}
&\mathcal{H}({\bm{s}}_i^{+}(t)) = -r \log(r) - (1-r) \log(1-r), \\
&\mathcal{H}({\bm{s}}_i^{\pm}(t)) = -2 \times \frac{r}{2}\log(\frac{r}{2}) - (1-r) \log(1-r).
\end{split}
\end{equation}
According to Eq.\ref{ternary1},\ref{ternary2}, information entropy achieves maximum, as long as the positive and negative spikes have the same probability.

\begin{figure}[t]
\begin{center}
\centerline{\includegraphics[width=\columnwidth]{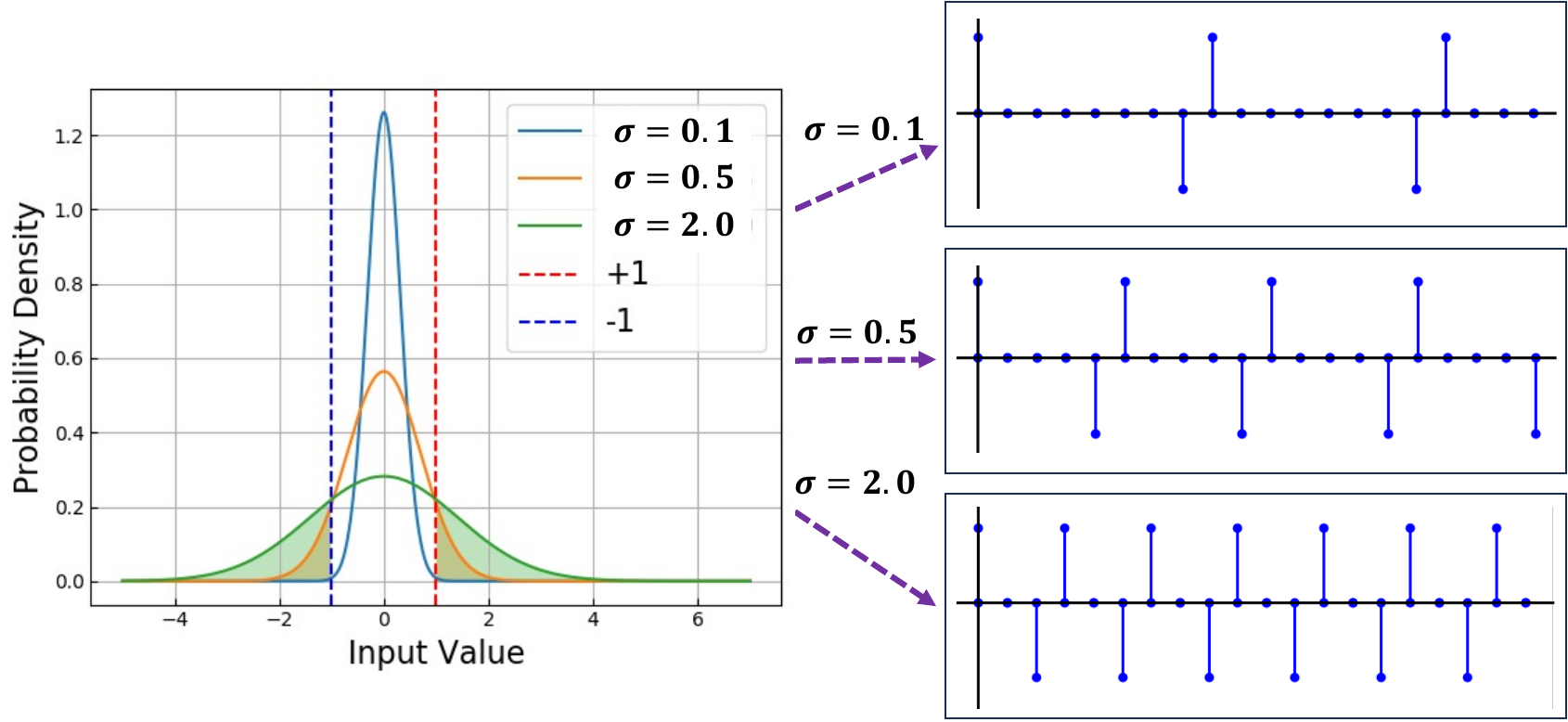}}
\caption{Relationship between the variance of input distributions and the spike firing frequencies.}
\label{freq}
\end{center}
\vskip -0.2in
\end{figure}

\subsection{Spike Frequency Encoding}
As shown in Eq.\ref{spike2}, previous spike encoding disregards input distributions, directly using a fixed threshold $\bm{\theta}^l$ to binarize inputs. 
However, the input distributions vary across different neurons. 
A key issue with previous spike encoding is its failure to perceive the variance of input distributions, resulting in difficulties to maintain a reasonable firing rate.

\textbf{Distribution-aware frequency encoding.}
We introduce a distribution-aware frequency encoding, which adjusts the input distribution for each neural layer. 
We achieve stable and manually controllable spike firing frequencies by elasticating membrane potentials with a scaling factor $\bm{\alpha}(t)$, and the elastic membrane potential is $\hat{\bm{m}}(t) = \bm{m}(t)/\bm{\alpha}(t)$.
As illustrated in Fig. \ref{freq}, by adjusting the variance of the input distribution, we can encode spikes in different frequencies, where a larger variance distribution results in a higher spike firing rate. We define the scaling factor as $k$ times the mean of membrane potential amplitude, $\frac{1}{n} \sum_{i=1}^{n} \left| \bm{m}_i(t) \right|$:
\begin{equation}
\label{alpha}
\bm{\alpha}^l(t) \stackrel{\mathrm{def}}{=} \frac{k}{n} \sum_{i=1}^{n} \left| \bm{m}_i^{l,(1)}(t) \right|,
\end{equation}
where $\bm{m}_i^{l,(1)}(t)$ indicates the membrane potential under the first batch of training data.
For stable training, we determine every $\bm{\alpha}^l(t)$ by the first batch, and then freeze $\bm{\alpha}^l(t)$ in training.
We replace the $\bm{m}^{l}(t)$ in Eq.\ref{certain1} with the elastic membrane potential $\hat{\bm{m}}^l(t)$ and obtain the frequency encoding:
\vspace{-0.2cm}
\begin{equation}
\label{alpha2}
{\bm{s}}^{\pm}(t) = \begin{cases} -1, & \text { if } \bm{m}(t)<-\bm{\alpha}(t) \\ 0, & \text { if } \bm{m}(t) \in (-\bm{\alpha}(t), +\bm{\alpha}(t)) \\ +1, & \text { if } \bm{m}(t) > +\bm{\alpha}(t) \end{cases}. 
\end{equation}
In Eq.\ref{alpha}, we define $k$ as an adjustable hyperparameter. When reducing $k$, it equally reduces the spike threshold in Eq.\ref{alpha2}, leading to an increased spike frequency; conversely, increasing $k$ reduces the spike frequency. 

Then, we will understand $\bm{\alpha}^l(t)$ from the perspective of the variance of input, which distribution is widely believed as roughly zero-mean Gaussian or Laplacian \cite{lin2022siman, banner2019post}. For simplification, we also use this assumption; for more complex distributions, it can be proved similarly as follows.

\textbf{Lemma 1.} \textit{Given a zero-mean Gaussion or Laplacian membrane potential m, $i.e., m\sim \mathcal{N}(0, \sigma^2)$ or $m \sim \text{La}(0, b)$, the scaling factor $\bm{\alpha}^l(t)$ is $\sqrt{\frac{2}{\pi}} \sigma k$ or $b k$.}

This is proved by calculating the expectation of ${\alpha}^l(t)$, and $\bm{\alpha}^l(t) = k\mathbb{E}[|\bm{m}^l(t)|] = k\int_{-\infty}^{\infty} |m| f(m) \, dm$, where $f(.)$ is zero-mean Gaussian or Laplacian distribution. 
As \textit{Lemma 1}, $\bm{\alpha}^l(t)$ is linearly related to the standard deviation $\sigma$.

\subsection{Spike Amplitude Encoding}
Eq.\ref{spike2} also neglects the intensity of membrane potential intensity. 
To preserve the intensity information, we encode the expectation of membrane potentials into spikes amplitude. 
This is achieved by scaling spike amplitude to $\bm{\alpha}^l(t)$, which formulates an identity transformation for membrane potentials in expectation:
\begin{equation}
{\bm{s}}^{\pm}(t) = \begin{cases} -\bm{\alpha}(t), & \text { if } \bm{m}(t)<-\bm{\alpha}(t) \\ 0, & \text { if } \bm{m}(t) \in (-\bm{\alpha}(t), +\bm{\alpha}(t)) \\ +\bm{\alpha}(t), & \text { if } \bm{m}(t) > +\bm{\alpha}(t) \end{cases}.
\label{b2}
\end{equation}
In this case, the backpropagation is the same as Eq.\ref{back}, since Eq.\ref{b2} is equivalent to dividing and then multiplying Eq.\ref{certain1} by $\bm{\alpha}(t)$ before and after respectively.
Due to the spike amplitude becoming $\bm{\alpha}^l(t)$, we accordingly revise the membrane potential update formula Eq.\ref{spike3} as:
\begin{equation}
\bm{v}^l(t)=\bm{m}^l(t)(\bm{\alpha}(t) - \bm{s}^{l}(t)) + {v}_{reset}\bm{s}^{l}(t).
\label{b3}
\end{equation}
Notably, at each time step $t$, we employ a layerwise amplitude encoding. Due to the commutative property of multiplication, we can first conduct matrix multiplications with unit amplitude spikes, and then reweight the results using spike amplitude. After training, the spike amplitude can merge with the weights in this layer. Amplitude encoding does not change the addition property of spike-driven operations.

\subsection{Conquering Language Modeling with SNNs}
We refer to the proposed spike encoding in Eq.\ref{spike1},\ref{b2},\ref{b3} as elastic bi-spiking mechanisms, which jointly encodes extended direction, frequency, and amplitude information of membrane potentials. 
We replace the traditional LIF neurons with elastic bi-spiking mechanisms and construct directly trainable language-oriented SNNs, termed SpikeLM.

We theoretically prove that elastic bi-spiking mechanisms ensure high optimization stability in SpikeLM, guaranteeing its performance in general language tasks. 
This is achieved by dynamical isometry: if a neural network achieves dynamical isometry, it prevents gradients from vanishing or exploding, maintaining nearly all values of its input-output Jacobian matrixes around one. 
A neural network can generally be viewed as a series of blocks $f^j_{\boldsymbol{\theta}^j}$ with parameters $\boldsymbol{\theta}^j$:
\begin{equation}
f(x_0)=f^L_{\boldsymbol{\theta}^L}\circ f^{L-1}_{\boldsymbol{\theta}^{L-1}} \circ \cdots \circ f^1_{\boldsymbol{\theta}^1}(x_0),
\label{eq:serial}
\end{equation}
where the Jacobian matrix $\frac{\partial f^{j}}{\partial f^{j-1}}$ is $\boldsymbol{J}_j$.  
It can be defined $\phi(\boldsymbol{J}) \stackrel{\mathrm{def}}{=} \mathbb{E}[tr(\boldsymbol{J})]$, and $\varphi(\boldsymbol{J}) \stackrel{\mathrm{def}}{=} \phi(\boldsymbol{J}^2)-\phi(\boldsymbol{J})^2$.

\textbf{Definition 1.} Block Dynamical Isometry (Definition 3.1 in \cite{chen2020comprehensive}).
\textit{Consider a neural network that can be represented as Eq.~\ref{eq:serial} and the $j$-th block’s Jacobian matrix is denoted as $\boldsymbol{J}_j$. If $ \forall$ j, $\phi(\boldsymbol{J}_j {\boldsymbol{J}^T_j})$ $\approx 1$ and $\varphi(\boldsymbol{J}_j {\boldsymbol{J}^T_j}) \approx 0$, the network achieves block dynamical isometry.}

\textbf{Lemma 2.} \textit{Given the probability of the input greater than 0 is $p$, the values of $\phi(\boldsymbol{J})$ and $\varphi(\boldsymbol{J})$ are $p$ and $p - p^2$ for the ReLU function.} (Proof in A.6 \cite{chen2020comprehensive})

\textbf{Lemma 3.} \textit{Given the spike fire rate is $r$, the values of $\phi(\boldsymbol{J})$ and $\varphi(\boldsymbol{J})$ are $1-r$ and $r - r^2$ respectively for the elastic bi-spiking function in Eq.\ref{b2}.}
\begin{proof} 
For clarity, we denote the elastic spike encoding (Eq.\ref{b2}) as $\bm{s}(\bm{m})$, where $\bm{m}$ is the membrane potential, and donate the Jacobian matrix as $\bm{s_m}$. Because $\bm{s}(\bm{m})$ is the element-wise operation, $\bm{s_m}$ is a diagonal matrix. According to Eq.\ref{back}, the gradient of $\bm{s}(\bm{m})$ is the STE between -1 and +1, so that, the value of $\bm{s_m}$ is 0 or 1. Given the spike firing rate $r$, the probability in [-1,+1] is $1-r$. Therefore, the spectral density of $\bm{s_m}$ is: $\rho_{\mathbf{s_m}}(z) = r\delta(z)+(1-r)\delta(z-1)$. And we have $\rho_{\mathbf{s_ms_m}^T}(z) = \rho_{\mathbf{s_m}}(z)$ because of the \{0,1\} matrix value. Accordingly, we have:
\begin{equation}
   \begin{split}
       &\phi(\mathbf{s_ms_m}^T) = \int_{\mathbb{R}}z \rho_{\mathbf{s_ms_m}^T}(z) dz = 1-r,\\
       &\varphi(\mathbf{s_ms_m}^T) = \int_{\mathbb{R}}z^2 \rho_{\mathbf{s_ms_m}^T}(z) dz - \phi^2(\mathbf{s_ms_m}^T) \\
       & ~~~~~~~~~~~~~~~~~~ = r - r^2.
   \end{split}
\label{equ:eig_s}
\end{equation}
\vskip -0.4in
\end{proof}

\textbf{Theorem 1.} \textit{In a deep neural network, the elastic bi-spiking function achieves better dynamical isometry than the ReLU: $\phi(\mathbf{s_ms_m}^T) > \phi(\mathbf{f_xf_x}^T)$, $\varphi(\mathbf{s_ms_m}^T) < \varphi(\mathbf{f_xf_x}^T)$.}
\begin{proof} 
In \textit{Lemma 2} and \textit{Lemma 3}, $p$ is usually believed 0.5 for zero-mean input distribution and $r$ is roughly 0.1 to 0.3 in SNNs. Accordingly, $\phi(\mathbf{s_ms_m}^T) > \phi(\mathbf{f_xf_x}^T)$ is achieved. Moreover, the function $f(x)=x-x^2$ achieves maximum given $x=0.5$, so that, $\varphi(\mathbf{s_ms_m}^T) < \varphi(\mathbf{f_xf_x}^T)$ is achieved.
\end{proof}

Based on \textit{Theorem 1}, the Jacobian matrix of Eq.\ref{b2} is closer to $\mathbf{I}$ than the ReLU function in ANNs. As a result, the elastic bi-spiking function has better optimization stability than ReLU at least. The training stability of SpikeLM is confirmed accordingly.

\section{Experiments}

\begin{table*}[t]
\vskip -0.1in
\begin{center}
\begin{small}
\centering
\caption{Comparisons between SpikeLM and ANNs or other SNNs on the GLUE dev set. The energy consumption is evaluated by FP32 operations and more results about FP16 energy consumption are in Table \ref{flops}. We report the evaluation metric in Appendix A.3. We implement LIF-BERT$^*$ and PSN-BERT$^*$ with LIF and PSN spiking neurons in Spikingjelly \cite{fang2023spikingjelly}, where the hyperparameters of spiking neurons follow previous SpikeBERT \cite{lv2023spikebert}. BERT$_\texttt{3L}$ and SpikeLM$_\texttt{6L}$ indicate the 3-layer BERT model and 6-layer SpikeLM respectively. We transfer the vision-oriented SpikingFormer \cite{zhou2023spikingformer} to language tasks by removing spiking neurons in the query and attention weight for a fair comparison with LIF-BERT and SpikeLM. Except for SpikeBERT \cite{lv2023spikebert}, all spiking models apply the same training scheme.}
\label{glue}
\setlength{\tabcolsep}{1.7mm}
\fontsize{9.3pt}{\baselineskip}\selectfont
\begin{tabular}{l|ll|cccccccc|c}
\midrule
\textbf{Model} & \textbf{Energy}$_\text{ (mJ)}$ & \textbf{Time}  & \textbf{MNLI}$_\text{-m/mm}$ & \textbf{QQP}$_\text{F1}$ & \textbf{QNLI} & \textbf{SST-2} & \textbf{CoLA} & \textbf{STS-B} & \textbf{MRPC}$_\text{F1}$ & \textbf{RTE} & \textbf{Avg.} \\ 
\midrule
BERT$_\texttt{base}$  & 51.41 & --   & 83.8/83.4 & 90.5 & 90.7 & 92.3 & 60.0 & 89.4 & 89.8 & 69.3 & 83.2 \\
BERT$_\texttt{3L}$    & 12.9  & --   & 77.1/77.1 & 85.2 & 85.8 & 88.1 & 31.7 & 85.7 & 86.4 & 66.4 & 75.9 \\
Q2BERT                & --    & --   & 47.2/47.3 & 67.0 & 61.3 & 80.6 & 0.0  & 4.7  & 81.2 & 52.7 & 49.1 \\
ELMo                  & --    & --   & 68.6/--   & 86.2 & 71.1 & 91.5 & 44.1 & 70.4 & 76.6 & 53.4 & 70.2 \\
\midrule
SpikeBERT             & 14.30 & 4    & 71.4/71.0 & 68.2 & 66.4 & 85.4 & 16.9 & 18.7 & 82.0 & 57.5 & 59.7 \\
LIF-BERT$^*$          & --    & 4    & 35.4/35.2 & 0.0  & 50.5 & 50.9 & 0.0  & 0.0  & 81.2 & 52.7 & 34.6 \\
PSN-BERT$^*$          & --    & 4    & 35.4/35.2 & 0.0  & 50.5 & 50.9 & 0.0  & 6.8  & 81.2 & 52.7 & 34.7 \\
LIF-BERT              & 7.98  & 4    & 56.8/55.2 & 70.0 & 60.6 & 80.6 & 14.6 & 20.0 & 82.3 & 53.8 & 54.9 \\
SpikingFormer         & --    & 1    & 67.8/68.6 & 79.3 & 74.6 & 82.7 & 16.7 & 72.3 & 83.0 & 58.8 & 67.1 \\
SpikingFormer         & --    & 4    & 70.2/70.6 & 80.9 & 79.5 & 83.9 & 12.8 & 77.0 & 83.0 & 62.1 & 68.9 \\
\midrule
SpikeLM$_\texttt{6L}$ & 2.05  & 1    & 73.9/75.3 & 83.2 & 84.2 & 86.2 & 30.7 & 83.7 & 85.7 & 66.8 & 74.4 \\
SpikeLM$_\texttt{6L}$ & 7.06  & 4    & 75.1/75.3 & 83.5 & 84.6 & 87.4 & 33.7 & 84.5 & 86.5 & 64.3 & 75.0 \\
SpikeLM               & 3.98  & 1    & 76.0/76.9 & 84.0 & 84.9 & 86.5 & 37.9 & 84.3 & 85.6 & 65.3 & 75.7 \\
SpikeLM               & 13.74 & 4    & 77.1/77.2 & 83.9 & 85.3 & 87.0 & 38.8 & 84.9 & 85.7 & 69.0 & 76.5 \\
\toprule
\end{tabular}
\end{small}
\end{center}
\vskip -0.2in
\end{table*}

\begin{table*}[h]
\begin{center}
\begin{small}
\centering
\caption{Comparisons between SpikeLM, SpikingFormer \cite{zhou2023spikingformer}, and ultra-low bit quantization methods on the GLUE dev set. Weight and Act. indicate the bit-width of weights and activations respectively, where ter indicates the ternary quantization level. For SpikeLM and SpikingFormer, we set the time step as 1. The 1-bit weight SpikeLM has similar operations with binary BERTs, which is because the binary BERTs have about 0.5$\times$ equivalent sparsity (Fig. \ref{fire0}) while SpikeLM has less activation firing rate in the BERT architecture.}
\label{glue2}
\setlength{\tabcolsep}{1.7mm}
\fontsize{9.3pt}{\baselineskip}\selectfont
\begin{tabular}{l|ll|cccccccc|c}
\midrule
\textbf{Model} & \textbf{Weight} & \textbf{Act.}  & \textbf{MNLI}$_\text{-m/mm}$ & \textbf{QQP}$_\text{acc}$ & \textbf{QNLI} & \textbf{SST-2} & \textbf{CoLA} & \textbf{STS-B} & \textbf{MRPC}$_\text{acc}$ & \textbf{RTE} & \textbf{Avg.} \\ 
\midrule
Q2BERT              & 2   & 8   & 47.2/47.3 & 67.0 & 61.3 & 80.6 & 0.0  & 4.4  & 68.4 & 52.7 & 47.7 \\
TernaryBERT         & Ter & Ter & 40.3/40.0 & 63.1 & 50.0 & 80.7 & 0.0  & 12.4 & 68.3 & 54.5 & 45.5 \\
BinaryBERT          & 1   & 1   & 62.7/63.9 & 79.9 & 52.6 & 82.5 & 14.6 & 6.5  & 68.3 & 52.7 & 53.7 \\
BiBERT              & 1   & 1   & 66.1/67.5 & 84.8 & 72.6 & 88.7 & 25.4 & 33.6 & 72.5 & 57.4 & 63.2 \\
BiT                 & 1   & 1   & 77.1/77.5 & 82.9 & 85.7 & 87.7 & 25.1 & 71.1 & 79.7 & 58.8 & 71.0 \\
BiPFT               & 1   & 1   & 69.5/70.6 & 83.7 & 81.7 & 86.2 & 22.9 & 80.2 & 76.2 & 66.1 & 70.8 \\
\midrule
SpikingFormer       & 32  & 1   & 67.8/68.6 & 83.8 & 74.6 & 82.7 & 16.7 & 72.3 & 74.0 & 58.8 & 66.6 \\
SpikeLM             & 32  & Ter & 76.0/76.9 & 87.9 & 84.9 & 86.5 & 37.9 & 84.3 & 78.7 & 65.3 & 75.4 \\
SpikeLM             & 1   & Ter & 74.9/75.1 & 87.2 & 84.5 & 86.6 & 36.0 & 83.9 & 78.9 & 65.7 & 74.8 \\
\toprule
\end{tabular}
\end{small}
\end{center}
\vskip -0.2in
\end{table*}

\begin{table}[t]
\vskip -0.1in
\caption{Energy consumption of the ANN-based BERT, LIF-BERT, and SpikeLM. We evaluate energy with both the FP16 and FP32 Multiply-ACcumulate (MAC) or ACcumulate (AC) operations.}
\label{flops}
\begin{center}
\begin{small}
\resizebox{\columnwidth}{!}{
\begin{tabular}{l|c|ccc}
\toprule
Model            & BERT$_\texttt{base}$ & LIF-BERT & SpikeLM & SpikeLM \\
\midrule
Steps                        & 1        &  4       &  1      &  4      \\
FP16$_\text{ (mJ)}$          & 15.21    &  3.55    &  1.77   &  6.10   \\
FP32$_\text{ (mJ)}$          & 51.41    &  7.98    &  3.98   &  13.74  \\
\bottomrule
\end{tabular}}
\end{small}
\end{center}
\vskip -0.3in
\end{table}

We evaluate previous SNNs and SpikeLMs on a range of general language tasks, including discriminative and generative. We mainly explore three key issues: \textbf{(i)} the baseline performance of traditional SNNs in general language tasks; \textbf{(ii)} the effectiveness of elastic bidirectional spike encoding in SpikeLM; and \textbf{(iii)} how to achieve controllable spike firing rate for energy efficiency.

\subsection{Settings}
\textbf{Language tasks.} For discriminative tasks, we evaluate SNNs on the standard GLUE benchmark\cite{wang2018glue}, which includes 8 subsets for classification and regression in different scenes. 
For generative tasks, we evaluate text summarization benchmarks: XSUM \cite{narayan2018don} and CNN-DailyMail \cite{nallapati2016abstractive}. 
Additionally, we evaluate the machine translation task on the WMT16 English-Romanian dataset \cite{bojar2016findings}.

\textbf{Architectures.} We develop SNN baselines and SpikeLM for discriminative and generative tasks using BERT and BART architectures respectively. For frequency encoding, we set $k=2$. As Section 3.1, we implement SNNs by replacing all matrix multiplications in ANNs with spike operations, maintaining the same architectures. Specifically, we use: \textbf{(i)} the BERT base \cite{devlin2018bert} for discriminative tasks, which is a 12-layer encoder transformer with 110M parameters; \textbf{(ii)} the BART base \cite{lewis2019bart} for text summarization, which is a encoder-decoder transformer with 6 layers for each and totally 139M parameters; and \textbf{(iii)} the mBART large model \cite{liu2020multilingual} for translation tasks, which is pretrained on 25 languages and has 680M parameters.

\subsection{Discriminative Tasks}
We follow the standard ANN-based BERT to develop SNN-based LIF-BERT and SpikeLM, which include two stages: pretraining and finetuning.
In pretraining, we use the BooksCorpus \cite{zhu2015aligning} and English Wikipedia \cite{devlin2018bert} as training data, including 800M and 2500M words respectively. 
In finetuning, we use the GLUE benchmark training with the common settings of ANNs. Training details are reported in Appendix A.2.

\textbf{Results of GLUE benchmark.}
As shown in Table \ref{glue}, we compare SpikeLM with both ANNs and SNNs. 
Our ANN baselines include BERTs \cite{devlin2018bert}, ELMo \cite{peters-etal-2018-deep}, and Q2BERT with 2-bit weights and 8-bit activations \cite{zhang2020ternarybert}, while the SNN baselines include SpikeBERT \cite{lv2023spikebert} and directly training SpikeingFormer \cite{zhou2023spikingformer}.
We additionally implement spike-driven BERTs with the PSN \cite{fang2023parallel} and LIF \cite{gerstner2014neuronal} neurons with original neuron settings \cite{fang2023spikingjelly,lv2023spikebert}. The difference between LIF-BERT and LIF-BERT$^*$ is in Appendix A.1.
Compared with BERT$_\texttt{base}$, SpikeLM reduces the performance gap to 6.7\%, while the original gap is 28.3\% in the LIF-BERT baseline.
We compare SpikeLM with SpikeBERT \cite{lv2023spikebert}, which is distilled from ANN-based BERT. SpikeLM exceeds 16.8\% average performance without any distillation, indicating the overall improvements in stand-alone learning capabilities of SNNs. Compared with original LIF-BERT$^*$ and PSN-BERT$^*$, SpikeLM dramatically improves 41.9\% and 41.8\% respectively. Results show that previous \{0,1\} spikes can not successfully model standard discriminative tasks. By leveraging elastic bi-spike encoding, their performance drop is effectively addressed.

\begin{figure}[t]
\begin{center}
\centerline{\includegraphics[width=0.95\columnwidth]{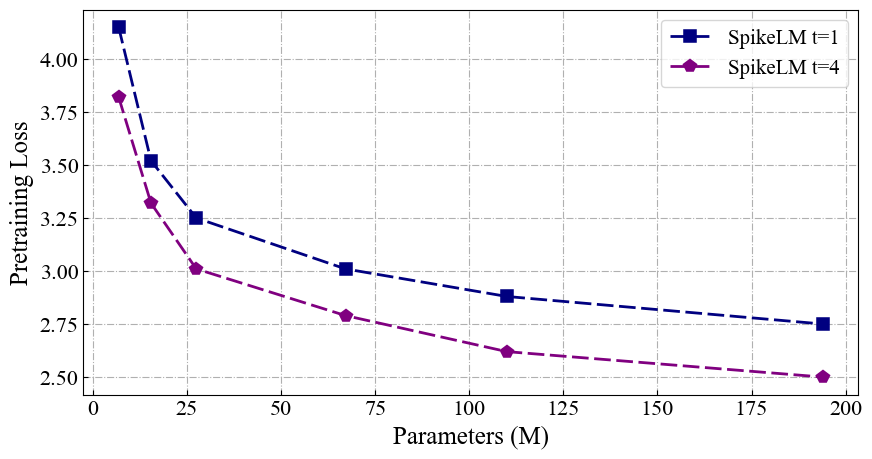}}
\caption{SNN scaling law of SpikeLM (T=1,4).}
\label{scale}
\end{center}
\vskip -0.4in
\end{figure}

\begin{figure*}[h]
\begin{center}
\centerline{\includegraphics[width=0.95\textwidth]{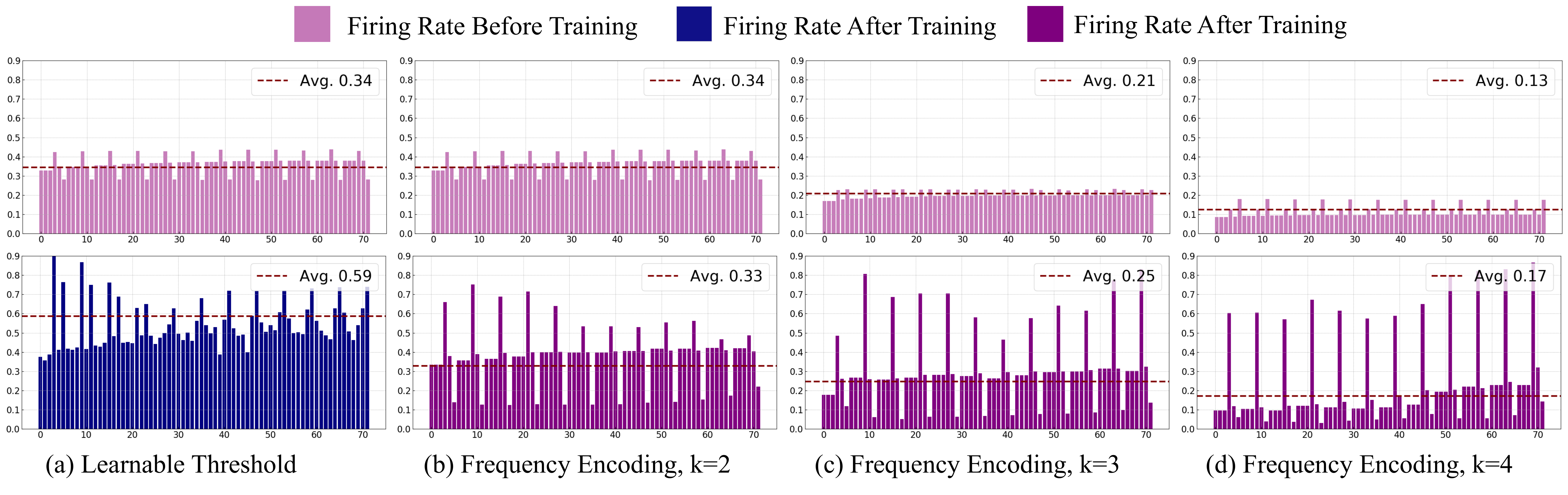}}
\caption{Spiking firing rate in linear layers under 2 settings: the learnable thresholds $\bm{\alpha}(t)$ (a) and spike frequency encoding (b, c, d).}
\label{fire}
\end{center}
\vskip -0.4in
\end{figure*}

As shown in Table \ref{glue2}, we also compare SpikeLMs with ultra-low bit quantization BERTs including Q2BERT \cite{zhang2020ternarybert}, TernaryBERT \cite{qin2022bibert, zhang2020ternarybert}, BinaryBERT \cite{qin2022bibert, bai2020binarybert}, BiBERT \cite{qin2022bibert}, BiT \cite{liu2022bit}, and BiPFT \cite{xing2024bipft}. Because of the sparse encoding in SpikeLM, the 1-bit weight SpikeLM (T=1) has similar operations to BERTs with both binary weights and activations. Specifically, we view the sparsity of BNNs as 0.5 according to Fig. \ref{fire0}, because value levels are able to map to \{0,1\} in inference. Compared with binary BERTs, SpikeLM also achieves higher performance.

\textbf{Energy efficiency.} As Table \ref{glue}, compared with BERT$_\texttt{base}$, SpikeLM saves 12.9$\times$ and 3.7$\times$ energy consumption with spike time steps 1 and 4 respectively.
Compared with SpikeBERT \cite{lv2023spikebert}, SpikeLM (T=1) exceeds 16.0\% average performance and also saves 3.6$\times$ energy. 
In Table \ref{flops}, it is shown that FP16 and FP32 operations have a similar tendency.

\begin{table}[t]
\caption{Ablation study on bidirectional spike encoding and spike frequency/amplitude encoding.}
\label{ablation}
\begin{center}
\begin{small}
\begin{tabular}{l|c|c}
\toprule
\textbf{Method} & \textbf{Spikes} & \textbf{GLUE Avg.} \\
\midrule
LIF-BERT$^*$ (T=4)             & $\{0,1\}$                 & 34.6  \\
\midrule
LIF-BERT (T=4)                 & $\{0,1\}$                 & 54.9  \\
+ Bidirect. encoding           & $\{-1,0,1\}$              & 55.7  \\
+ Freq./Amp. encoding          & $\{-\alpha,0,\alpha\}$    & 76.5  \\
\midrule
LIF-BERT (T=4)                 & $\{0,1\}$                 & 54.9  \\
+ Freq./Amp. encoding          & $\{0,\alpha\}$            & 71.8  \\
+ Bidirect. encoding           & $\{-\alpha,0,\alpha\}$    & 76.5  \\
\bottomrule
\end{tabular}
\end{small}
\end{center}
\vskip -0.2in
\end{table}

\textbf{SNN scaling law.} As shown in Fig. \ref{scale}, we explore the scalability of language-oriented SNNs by adjusting the parameter number with different model widths.
We pretrain SpikeLMs from 6.9M to 194M parameters and use pretraining loss, including the mask language modeling and next sentence prediction, as the evaluation metric. For larger models, Wikipedia and BooksCorpus may be insufficient for pretraining, and larger-scale datasets are needed. The experiments show that the elastic bi-spiking mechanism follows the scaling law, supporting SpikeLM's scalability to some extent.

\textbf{Ablation study.}
The improvements of SpikeLM are attributed to elastic bi-spiking mechanisms, by encoding the direction, frequency, and amplitude of spikes. Notably, the frequency and amplitude encoding are coupled in Eq. \ref{b2}. Therefore, we analyze the individual contributions of bidirectional and frequency/amplitude encoding in Table \ref{ablation}.
When implementing backpropagation by Straight-Through Estimator (STE), adding spike frequency/amplitude encoding and bidirectional spike encoding improves the GLUE performance by 16.9\% and 4.7\%, demonstrating the enhanced modeling capacity of elastic bi-spiking mechanism. As a result, our visualization in Appendix A.4 demonstrates the extended bidirectional and amplitude information in spikes, and a proper firing rate is maintained.

\begin{table}[t]
\caption{Trade-off between performance and efficiency. N/A indicates directly setting spike thresholds as -1 and +1 as Eq.\ref{certain1}.}
\label{trade}
\vskip -0.3in
\begin{center}
\begin{small}
\begin{tabular}{l|c|c|ccc}
\toprule
Settings & N/A & Learnable & k=2 & k=3 & k=4 \\
\midrule
GLUE Avg.$_\text{ (\%)}$   & 55.7 & 76.8   & \textbf{76.5} & 75.1 & 60.8 \\
Erengy$_\text{ (mJ)}$      & 17.3 &  23.8  & \textbf{13.7} & 10.4 & 7.4  \\
\bottomrule
\end{tabular}
\end{small}
\end{center}
\vskip -0.2in
\end{table}

\begin{table*}[t]
\vskip -0.1in
\label{generate1}
\centering
\begin{small}
\caption{Comparisons between SpikeLM and ANNs or LIF-BARTs on generative tasks. We evaluate text summarization on XSUM and CNN-DailyMail with rouge-\{1,2,L\} metrics; and evaluate translation on the WMT16 En-Ro dataset with the BLEU metric. For XSUM and CNN-DailyMail, we use the base BART architecture; for WMT16, we use the large mBART architecture. We evaluate average energy consumption on the XSUM dataset for the base-sized BART, LIF-BART, and SpikeLM. Following other SNN works \cite{zhou2023spikingformer,yao2023spike}, we also apply FP32 operations for energy evaluation.}
\begin{tabular}{p{3.1cm} | C{1.5cm}| C{1.2cm} C{1.2cm} C{1.2cm} | C{1.2cm} C{1.2cm} C{1.2cm} | C{1.5cm} } 
\toprule
& \multicolumn{1}{c|}{\textbf{XSUM}} & \multicolumn{3}{c|}{\textbf{XSUM}} & \multicolumn{3}{c|}{\textbf{CNN-DailyMail}} & \multicolumn{1}{c}{\textbf{WMT16}}   \\
\cmidrule(lr){1-1}
\cmidrule(lr){2-2}
\cmidrule(lr){3-5}
\cmidrule(lr){6-8}
\cmidrule(lr){9-9}
\textbf{Model}                       & \textbf{Energy}$_\text{ (mJ)}$ & Rouge-1 & Rouge-2 & Rouge-L & Rouge-1 & Rouge-2 & Rouge-L & BLEU \\
\midrule
BART$_\texttt{base}$/mBART$_\texttt{large}$ & 378.22 & 42.75  & 19.80  & 34.72  & 44.88  & 22.24  & 31.75  & 26.82 \\
BERTSUMABS                           & --     & 38.76  & 16.33  & 31.15  & 41.72  & 19.39  & 38.76  & --    \\
PTGEN                                & --     & 29.70  & 9.21   & 23.24  & 36.44  & 15.66  & 33.42  & --    \\
GPT-2                                & --     & --     & --     & --     & 29.34  & 8.27   & 26.58  & --    \\
\midrule
LIF-BART (T=1)                       & 6.59   & 33.17  & 12.30  & 26.56  & 39.88  & 16.77  & 26.15  & 14.72 \\
LIF-BART (T=4)                       & 28.92  & 35.47  & 13.84  & 28.32  & 41.39  & 18.33  & 28.05  & 19.04 \\
\midrule
SpikeLM (T=1)                        & 25.68  & 39.40  & 17.13  & 31.97  & 41.56  & 18.74  & 28.51  & 20.93 \\
SpikeLM (T=4)                        & 115.79 & 40.63  & 17.97  & 32.92  & 42.19  & 19.27  & 29.06  & 22.95 \\
\bottomrule
\end{tabular}
\end{small}
\vskip -0.1in
\end{table*}

\textbf{Controllable spike firing rate.}
As Section 3.2, a key issue with existing SNNs is the trade-off between energy and performance. 
Spike frequency encoding can achieve a controllable firing rate, thereby enabling a manageable balance between performance and energy consumption. To estimate this, we compare two settings: \textbf{(i)} setting $\bm{\alpha}(t)$ as learnable parameters in each spike layer, as shown in Fig.\ref{fire}(a), and  \textbf{(ii)} setting $\bm{\alpha}(t)$ as Eq.\ref{alpha} with $k=2,3,4$, as shown in Fig.\ref{fire}(b, c, d). In Fig.\ref{fire}, we compare the distributions of spike firing rate in each linear layer of SpikeLM BERT models. We have the following results:
\textbf{(i)} In Fig.\ref{fire}(a), the freely trainable $\bm{\alpha}(t)$ leads to excessively high spike firing rates due to maximizing the entropy of the multinomial distribution. 
\textbf{(ii)} Spike frequency can be effiectively controlled by hyperparameter $k$ in Eq.\ref{b2}. By increasing k from 2 to 4, the average firing rate changes from 33\% to 17\%. 
\textbf{(iii)} Spike frequency encoding controls the firing rates without much performance drop. Compared to learnable thresholds, the frequency encoding at $k=2$ has almost the same performance and saves 42.4\% of energy in Table \ref{trade}.
\textbf{(iv)} The firing rate remains similar before and after training with spike frequency encoding, allowing to predict firing rate at the beginning of training.

\subsection{Generative Tasks}
We evaluate fully spike-driven language models in generative tasks for the first time.
Generation tasks require extended input and output sequence lengths, which necessitates advanced language modeling capabilities in SNNs. 
Therefore, we introduce the distillation strategy, which involves initializing and employing knowledge distillation from a pretrained ANN teacher.
In detail, we select the pretrained BART-base and mBART-large models as ANN teachers for summarization and translation. 
Training details are shown in Appendix A.2.

\textbf{Results of summarization.} In Table 8, we compare SpikeLM with both ANN and SNN baselines. Compared with the BART base model, the ROUGE-L of SpikeLM (T=4) only drops 1.80\% and 2.69\% on XSUM and CNN-DailyMail, despite replacing all the matrix multiplications as spikes-driven. Compared with GPT-2 and other ANNs, SpikeLMs can be also competitive. This is the first time to verify fully spike-driven models achieve competitive performance with ANNs in challenging generative tasks.

Compared with LIF-BARTs on XSUM, SpikeLMs exceed 5.41\% with one-step spikes and 4.60\% with 4-step spikes. Similar results are also shown on CNN-DailyMail. 
This indicates the omnidirectionally improved bidirectional spikes with levels $\{-\alpha,0,\alpha\}$ achieve much more capabilities in language modeling than previous $\{0, 1\}$ spikes.

\textbf{Results of translation.} In Table 8, we go further to evaluate SpikeLMs on multilingual tasks and large-sized mBART architecture. 
We observe that, even with the large-sized model, the performance of LIF-BART (T=4) remains inferior to SpikeLM (T=1). Therefore, improving spike capacity by generalized encodings is more effective than increasing spike time steps in previous SNNs.

\section{Conclusion}
This work proposes a fully spiking mechanism for general language tasks, demonstrating the potential generalization capacity of SNNs at a higher level. 
Unlike previous binary spikes, spike capabilities are significantly extended from bi-direction, amplitude, and frequency encodings, while maintaining the addition nature of SNNs.
Inspired by advanced neuroscience, it would be great potential to develop efficient and environmentally friendly large language models with spike-driven methods in the future.


\section*{Limitations}

The limitations of this work include the activation spiking in traditional SNNs and the model scale. Compared with weight quantization, activation quantization in SNNs is more challenging. Moreover, recent large language models are memory-bounded, and leveraging spiking neuronal dynamics to quantize weights may achieve higher performance and efficiency in the future. 

\section*{Acknowledgements}

This work is supported by National Key R\&D Program of China 2022ZD0160602, Natural Science Foundation of China 62122088, National Science Foundation of China under grant No.62206150, Distinguished Young Scholars (62325603), National Natural Science Foundation of China (62236009,62441606), and Beijing Natural Science Foundation for Distinguished Young Scholars (JQ21015).

\section*{Impact Statement}

This paper presents work whose goal is to advance the field of Machine Learning. There are many potential societal consequences of our work, none of which we feel must be specifically highlighted here.


\bibliography{example_paper}
\bibliographystyle{icml2024}

\newpage
\appendix
\onecolumn

\section{Appendix}
\subsection{Implementation Details of the LIF-based Transformers}
For comparison, we implement Leaky integrate-and-fire (LIF) neurons \cite{gerstner2014neuronal} with 2 backpropagation algorithms: 
\begin{itemize}
\item the original arctangent-like surrogate gradient function. We implement LIF-BERT$^*$ by Spikingjelly \cite{fang2023spikingjelly}, which is a popular open-source SNN framework with previous spike neurons. 
\item To make a strict comparison with our method in Section 4, we also propose a straight-through estimator (STE) \cite{bengio2013estimating} based backpropagation for LIF neurons. We implement our LIF-BERT/BART baselines in this way by PyTorch. 
\end{itemize}

\textbf{Forward propagation.} The same as Eq.\ref{spike2}, the membrane potential is binarized by Eq.\ref{spike21}, the $\bm{\theta}^{l}$ is 1 in usual:
\begin{equation}
\label{spike21}
\bm{s}^{l}(t)= \begin{cases}0, & \text { if } \bm{m}^{l}(t) < \bm{\theta}^{l} \\ 1, & \text { if } \bm{m}^{l}(t) \ge \bm{\theta}^{l} \end{cases}.
\end{equation}
Both of our and the original implementations are the same in forward propagation. The difference is how to relax this non-differentiable function for gradient calculation. 

\textbf{Backward propagation.} In the original LIF neurons, a gradient surrogate function is used: 
\begin{equation} \label{equ:surrogate}
\begin{aligned}
\bm{s}^{l}(t) \approx \frac{1}{\pi} \arctan(\frac{\pi}{2}\alpha \bm{m}^{l}(t)) + \frac{1}{2}
\end{aligned},
\end{equation}
which is arctangent-like to simulate Eq.\ref{spike21}. The following gradient estimation is used accordingly:
\begin{equation} \label{equ:gradients}
\begin{split}
\frac{\partial \bm{s}^{l}(t)}{\partial \bm{m}^{l}(t)} & =\frac{\alpha}{2} \frac{1}{(1+(\frac{\pi}{2}\alpha \bm{m}^{l}(t))^{2})}
\end{split}.
\end{equation}

To make a comparison with our method in Section 4, we also implement a similar STE-based backpropagation and evaluate the performance. Similar to Section 4.1, we relax the LIF neuron output as stochastic variables $\widetilde{\bm{s}}(t)$:
\begin{equation}
\label{ternary11}
\widetilde{\bm{s}}(t) \stackrel{\mathrm{def}}{=} \begin{cases}0, & \bm{p}^{0} = {\rm clip}(1 - \bm{m}(t), 0, 1) \\ 1, & \bm{p}^{+} =  {\rm clip}(\bm{m}(t), 0, 1) \end{cases}, 
\end{equation}
where $\bm{p}^{+}$ and $\bm{p}^{0}$ indicate the probability of 1 and 0. We define the $\bm{p}^{+}$ and $\bm{p}^{0}$ according to their distance to 1 and 0, and the ${\rm clip}(.)$ operation confirms the probability in [0,1].
As a result, we can use the gradient expectation of the stochastic variable $\widetilde{\bm{s}}(t)$ for backpropagation, similar to Eq.\ref{back}:
\begin{equation}
\label{back1}
\begin{split}
& \mathbb{E}_{\widetilde{\bm{s}}(t)}[\frac{\partial \widetilde{\bm{s}}(t)}{\partial \bm{m}(t)}] = \frac{\partial}{\partial \bm{m}(t)} \mathbb{E}[\widetilde{\bm{s}}(t)] \\
& \ \ = \frac{\partial}{\partial \bm{m}(t)} (0 \times \bm{p}^{0} + 1 \times \bm{p}^{+}) \\
& \ \ = \frac{\partial}{\partial \bm{m}(t)} {\rm clip}(\bm{m}(t),0,1) \\
\end{split}.
\end{equation}
In our LIF-BERT/BART baselines, we use Eq.\ref{back1} in backward to compare the elastic bi-spiking mechanisms with Eq.\ref{back}. For the spike neurons in linear layers, both our LIF-BERT/BART baseline and SpikeLM set $\beta$ in Eq.\ref{spike3} as 0.25 the same as other works; for the spike neurons in the key-value cache of self-attention, the $\beta$ is set to 0, leading to not considering the results of last time step. This would confirm the parallelism of self-attention operations. For the original LIF-BERT$^*$ implementation, we apply the same hyperparameters of spiking neurons as the previous SpikeBERT \cite{lv2023spikebert}. As shown in Table \ref{glue1}, our implemented LIF-BERT baseline has higher accuracy.

\begin{table*}[t]
\begin{center}
\begin{small}
\centering
\caption{Comparisons between our implemented baseline LIF-BERT and the original LIF-BERT$^*$.}
\label{glue1}
\setlength{\tabcolsep}{1.7mm}
\fontsize{9.3pt}{\baselineskip}\selectfont
\begin{tabular}{l|c|cccccccc|c}
\midrule
\textbf{Model} & \textbf{Time}  & \textbf{MNLI}$_\text{-m/mm}$ & \textbf{QQP} & \textbf{QNLI} & \textbf{SST-2} & \textbf{CoLA} & \textbf{STS-B} & \textbf{MRPC} & \textbf{RTE} & \textbf{Avg.} \\ 
\midrule
LIF-BERT$^*$ (Original)          & 4    & 35.4/35.2  & 0.0  & 50.5 & 50.9 & 0.0  & 0.0  & 81.2 & 52.7 & 34.6 \\
LIF-BERT (Ours)                  & 4    & 56.8/55.2  & 70.0  & 60.6 & 80.6 & 14.6  & 20.0  & 82.3 & 53.8 & 54.9 \\
\toprule
\end{tabular}
\end{small}
\end{center}
\end{table*}

\subsection{Experiment Details}
\subsubsection{GLUE Benchmark}
In the pretraining phase, we keep the settings of the SNN baselines and SpikeLM similar to BERTs. 
As shown in Table \ref{glue}, our trained baselines include the original LIF-BERT$^*$, PSN-BERT$^*$, and our implemented LIF-BERT. 
We utilize the BooksCorpus \cite{zhu2015aligning} and English Wikipedia \cite{devlin2018bert} as our training datasets, which include 800M and 2500M words respectively. The same as the approach taken in BERT \cite{devlin2018bert}, lists, tables, and headers are ignored in Wikipedia.
In the preprocessing stage, our approach aligns with BERT's methodology, employing the WordPiece tokenizer \cite{devlin2018bert} with a 30522 vocabulary size. We set the maximum length of each sentence as 128 tokens. The batch size is set to 512 in training. The entire pretraining encompasses a total of $10^5$ steps.
The same as ANN conditions, we train SNNs with an AdamW optimizer with a $2\times10^{-4}$ peak learning rate and 0.01 weight decay.
We adapt the learning rate by a linear schedule with 5000 warm-up steps.
Our experiments show the commonly used pretraining hyperparameters for ANN-based BERTs are general and robust enough for SNNs.

We apply the standard GLUE benchmark \cite{wang2018glue} to evaluate the natural language understanding performance of LIF-BERT and SpikeLM. 
We follow previous works and use the 8 subsets, including CoLA, STS-B, MRPC, RTE, QQP, MNLI, and QNLI for classification or regression in different scenes. 
For evaluation, we follow BERT \cite{devlin2018bert} and report F1 scores for QQP and MRPC datasets; Spearman correlations for the STS-B dataset; and accuracy scores for other datasets. 
In the finetuning phase, we maintain commonly used hyperparameters for ANN-based BERTs. 
Specifically, we maintain a constant learning rate of $2\times10^{-5}$ and a batch size of 32 for all subsets. We keep the same training epochs as previous BiPFTs \cite{xing2024bipft} for different datasets. 
It's important to note that we do not adapt to the best learning rate or batchsize for GLUE subsets. This can improve performance a lot but may potentially overestimate performance when applied to new tasks.

\subsubsection{Generative Benchmarks}
For generative tasks, we use the XSUM and CNN-DailyMail summarization benchmarks, and the WMT16 En-Ro dataset as a translation benchmark.
XSUM is sampled from the BBC news website, including 226k documents and their one-sentence summarizations. 
CNN-DailyMail has longer documents and multi-sentence summarizations, and there are 300k data pairs.

For XSUM, CNN-DailyMail, and WMT16 datasets, we use the AdamW optimizer and train 20 epochs with a 128 batch size, and a peak learning rate of $3.5\times10^{-4}$, $7\times10^{-4}$, or $1\times10^{-4}$ respectively. 
We adapt the learning rate by a linear schedular with 0.05$\times$ total steps' warm-up. 
All SNN models are trained on a single node with 8 A800 GPUs. 

We distill the SNN-based BART models following the model compression methods \cite{li2022dq}.
As training objectives, we jointly use the cross-entropy loss and additionally the distillation loss including K-L divergence for the last-layer logits, and L2 loss for the hidden states and attention map in each layer.

\subsection{Energy Consumption Metric}
Following previous works, we evaluate the overall energy consumption of a neural network by the energy consumption of accumulate operations $E_{AC}$ and multiply-accumulate operations $E_{MAC}$. Under the 45nm process technology, the 32-bit floating point has an energy consumption of $E_{AC}=0.9pJ$ and $E_{MAC}=4.6pJ$ \cite{horowitz20141, zhang2022pokebnn}. Moreover, for FP16 operations, we apply the energy consumption of $E_{AC}=0.4pJ$ and $E_{MAC}=1.5pJ$ respectively.

For ANNs, the overall energy consumption can be directly evaluated by their MACs. For example, given a linear layer with input dimension $m$ and output dimension $n$, its energy consumption can be:
\begin{equation}
E_{Linear} = m \times n \times E_{MAC}.
\end{equation}

For SNNs, the energy consumption is also determined by the spike firing rate $r$ in a certain layer, and also the time step $T$ of the whole SNN. Given the same example as the ANN case, the energy consumption of the linear layer can be:
\begin{equation}
E_{Linear} = m \times n \times E_{AC} \times T \times r,
\end{equation}
because of the $r$ times sparsity in this spike neuron and T times calculation of the SNN. Different from multiply-accumulate operations (MACs) in ANNs, SNNs convert matrix multiplications to pure accumulate operations (ACs). In Table \ref{glue}, \ref{flops}, we evaluate energy by sampling the same 64 pretraining datas from the first batch. In Table 8, we evaluate energy on the XSUM test set.

\subsection{Visualization of Spike Neurons}
\begin{figure}[h]
\begin{center}
\centerline{\includegraphics[width=0.8\textwidth]{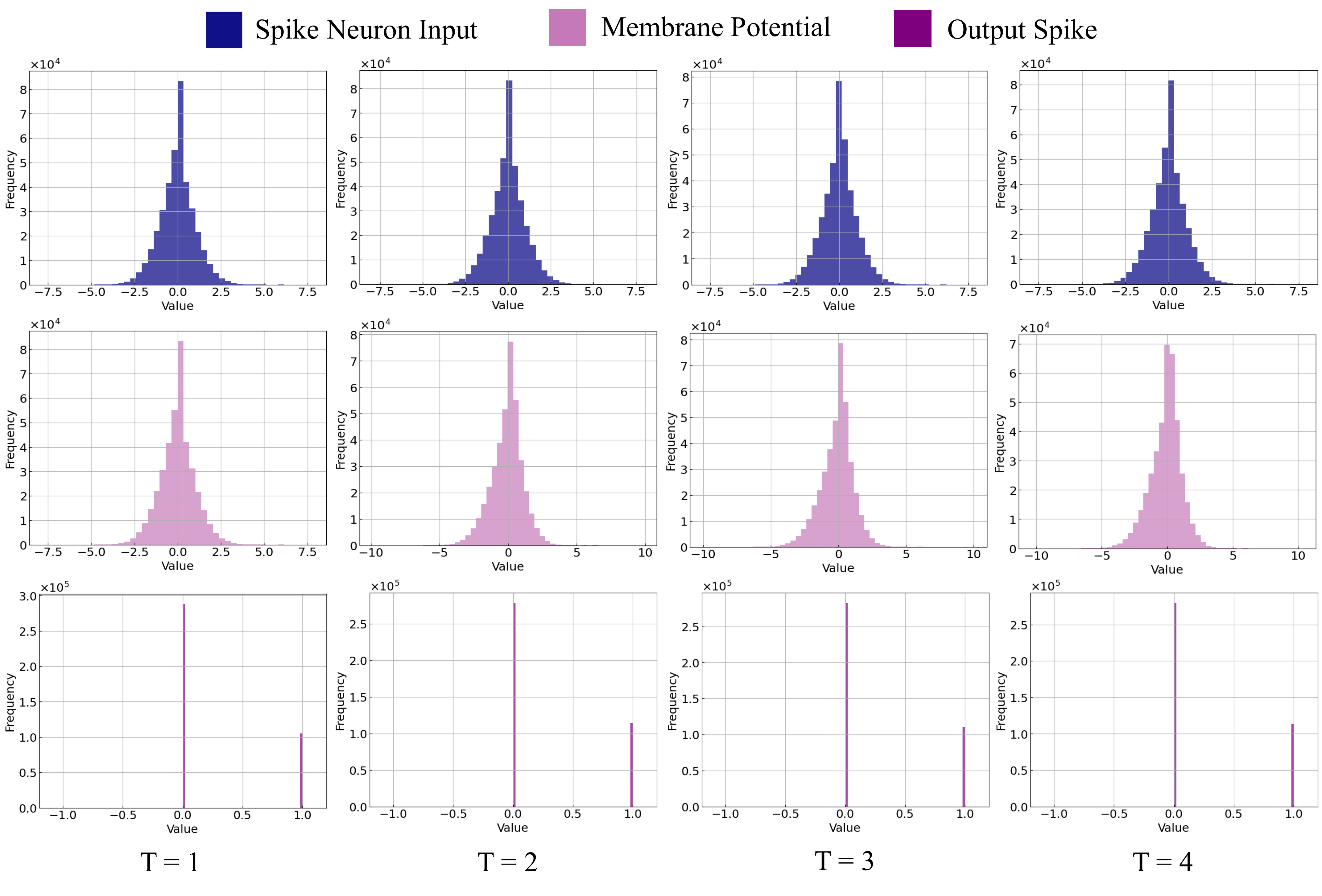}}
\caption{Visualization of the Leaky Integrate-and-Fire (LIF) neuron with $\{0,1\}$ spike levels.}
\label{view}
\end{center}
\end{figure}

\begin{figure}[h]
\begin{center}
\centerline{\includegraphics[width=0.8\textwidth]{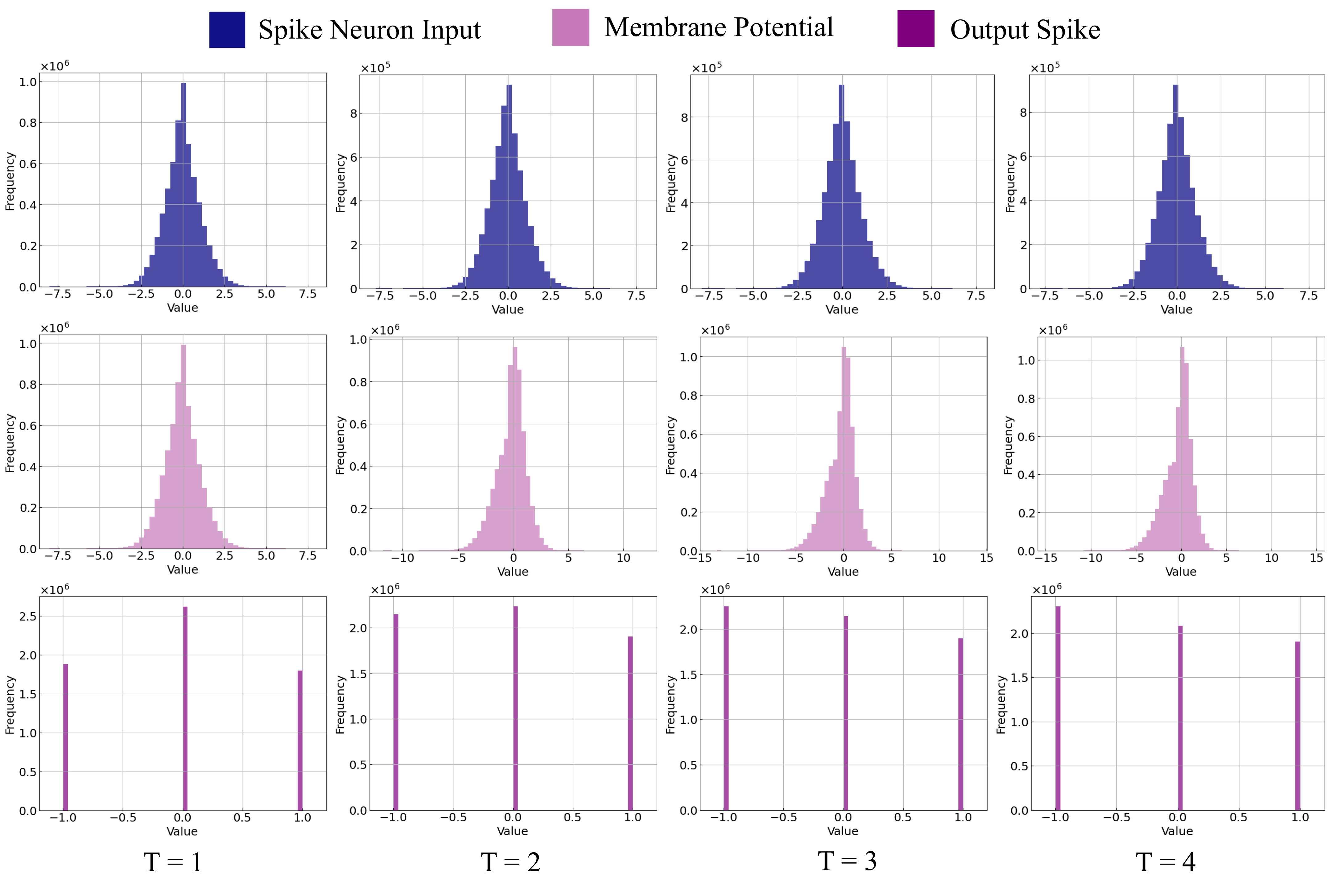}}
\caption{Visualization of the bidirectional spike encoding with $\{-1,0,1\}$ spike levels.}
\label{view2}
\end{center}
\end{figure}

\begin{figure}[h]
\begin{center}
\centerline{\includegraphics[width=0.8\textwidth]{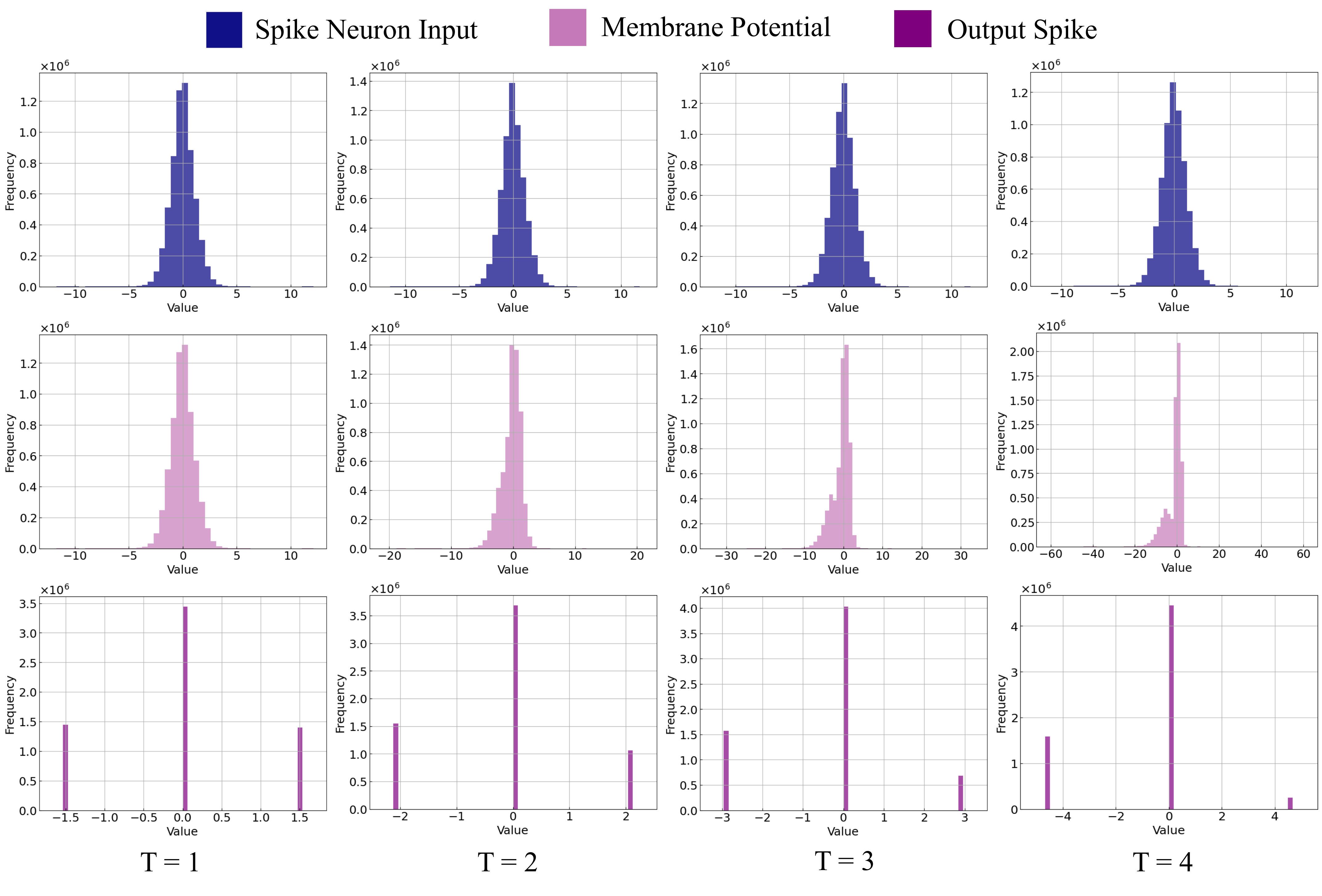}}
\caption{Visualization of the elastic bi-spiking mechanism with $\{-\alpha,0,\alpha\}$ spike levels.}
\label{view3}
\end{center}
\end{figure}

We compare three settings of spike neurons and visualize their input distillations, membrane potentials, and generated spikes in every SNN time step. For comparison, we select the first feed-forward linear layer in the first transformer block. We randomly sample 64 pretraining data of the BERT-architectured models and acquire the input, membrane potential, and output spike of the selected spike neurons.
\begin{itemize}
\item As shown in Fig. \ref{view}, we visualize the previous LIF neuron with binary spike levels $\{0,1\}$.
\item As shown in Fig. \ref{view2}, we visualize the LIF neuron with the bidirectional spike encoding proposed in Section 4.1. So that, the generated spikes have the ternary spike levels $\{-1,0,1\}$. However, this setting leads to a much higher spike firing rate, causing the energy consumption problem.
\item As shown in Fig. \ref{view3}, we visualize the elastic bi-spiking mechanism in SpikeLMs, which includes bidirectional spiking encoding, spike frequency encoding, and spike amplitude encoding. It is shown that the spikes not only have ternary levels $\{-\alpha,0,\alpha\}$, but also have the amplitude $\alpha$. In the frequency aspect, compared with Fig. \ref{view2}, the elastic bi-spiking mechanism achieves a lower spiking firing rate, demonstrating the effectiveness of spike frequency encoding. 
\end{itemize}

\end{document}